\pdfoutput=1

\documentclass[11pt]{article}

\usepackage{ACL2023}

\usepackage{times}
\usepackage{latexsym}
\usepackage{graphicx}
\usepackage{adjustbox}
\usepackage{algorithm}
\usepackage{algorithmic}
\usepackage{amsmath}
\usepackage{tablefootnote}

\usepackage[T1]{fontenc}

\usepackage[utf8]{inputenc}

\usepackage{microtype}

\usepackage{inconsolata}

\title{Attribute Controlled Dialogue Prompting}

\author{Runcheng Liu$^{1,2}$\thanks{\hspace{1mm} Work done during an internship at Huawei.}\hspace{1mm},
    Ahmad Rashid$^{1,2}$\samethanks \hspace{1mm},
    Ivan Kobyzev$^3$ \\
    {\bf Mehdi Rezagholizadeh$^{3}$,
    Pascal Poupart$^{1,2}$} \\
    $^1$David R. Cheriton School of Computer Science, University of Waterloo \\
    $^2$Vector Institute, Canada \\
    $^3$Huawei Noah’s Ark Lab, Canada \\
    \texttt{\{ireneliu,a9rashid,ppoupart\}@uwaterloo.ca} \\
    \texttt{\{ivan.kobyzev,mehdi.rezagholizadeh\}@huawei.com}}

\begin{document}
\maketitle
\begin{abstract}
Prompt-tuning has become an increasingly popular parameter-efficient method for adapting large pretrained language models to downstream tasks. However, both discrete prompting and continuous prompting assume fixed prompts for all data samples within a task, neglecting the fact that inputs vary greatly in some tasks such as open-domain dialogue generation. In this paper, we present a novel, instance-specific prompt-tuning algorithm for dialogue generation. Specifically, we generate prompts based on instance-level control code, rather than the conversation history, to explore their impact on controlled dialogue generation. Experiments on popular open-domain dialogue datasets, evaluated on both automated metrics and human evaluation, demonstrate that our method is superior to prompting baselines and comparable to fine-tuning with only 5\%-6\% of total parameters.
\end{abstract}

\section{Introduction}

Fine-tuning has been frequently used when deploying generative pretrained language models (PLMs) to downstream tasks since the advent of GPT~\cite{radford2018improving} and BERT~\cite{devlin2019bert}. However, this requires storing a full copy of parameter states for every downstream task, which is memory-consuming and expensive to serve when working with large-scale models with billions of parameters like GPT-3~\cite{brown2020language}.

In this work, we design a lightweight prompting module for adapting pretrained language models for attribute controlled dialogue generation. More precisely, for each attribute such as persona, intention, emotion etc. we only save an additional prompt module. Since the prompting module is a fraction of the size of the pretrained dialogue model, this allows many controlled dialogue systems to be stored on a device without too much overhead. We present results on both intent and persona controlled dialogue.

\begin{figure*}[htp]
    \centering
    \includegraphics[width=\textwidth]{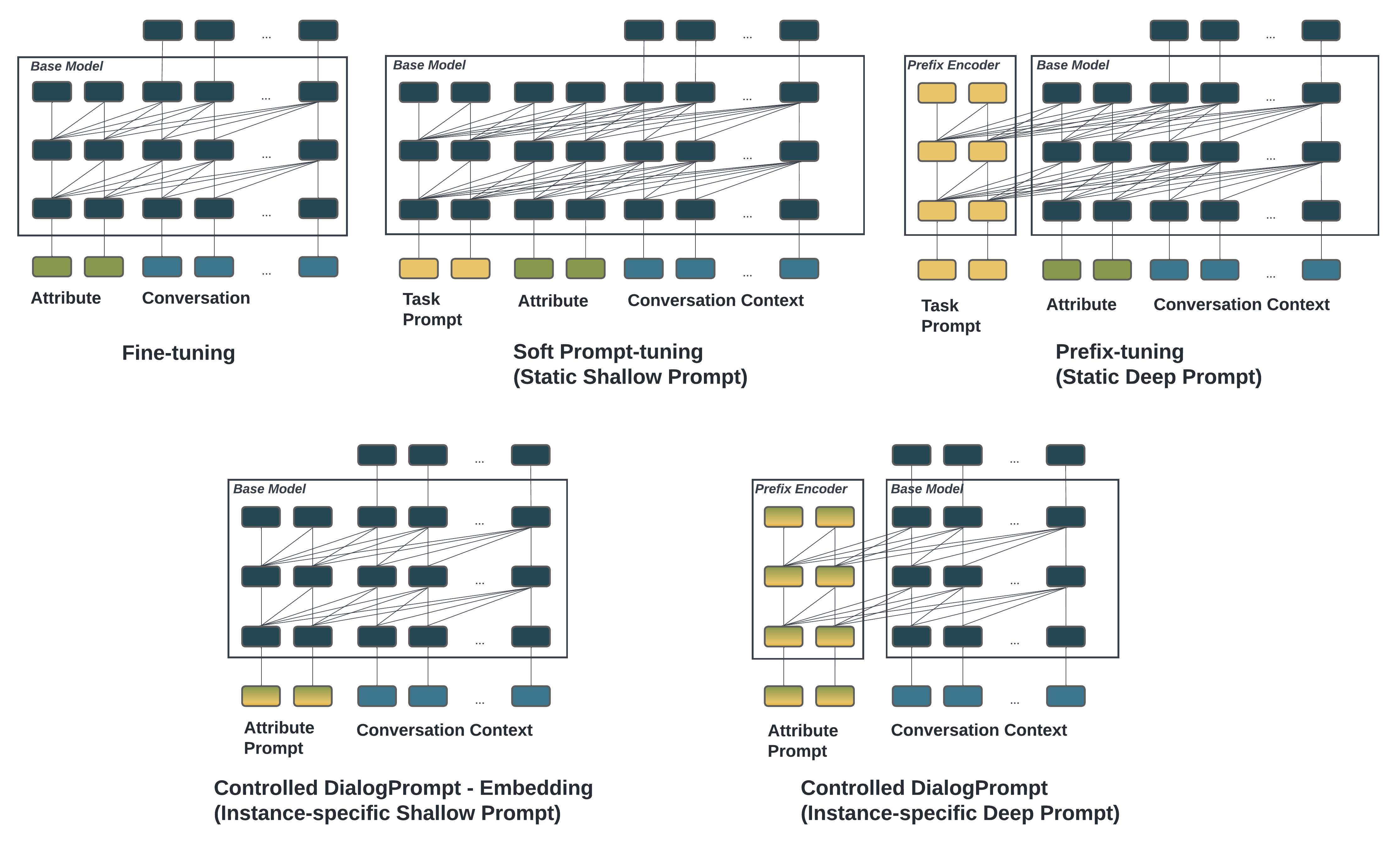}
    \caption{Diagrams illustrating attention mechanisms for different configurations of (task/attribute) prompt, attribute and conversation context.}
    \label{fig:ControlledDP}
\end{figure*}

\section{Related Work}

GPT-3~\cite{brown2020language} introduces {\em prompting}, a method to steer a frozen PLM by transforming inputs into cloze-style phrases with task description and some task examples. Though it is memory-efficient since one single copy of the PLM can be shared across different tasks, the model's performance is largely restricted by the maximum conditional input length, the model size and manual guesswork for prompts~\cite{zhao2021calibrate, schick2021exploiting, schick2021s, jiang2020can}. Other works focus on automatically searching for better discrete prompts~\cite{jiang2020can, shin2020autoprompt, gao2021making, ben2021pada}.

Recently, there has been an increased interest in {\em continuous prompts / prompt-tuning}, which bridges the gap between prompting and fine-tuning, while remaining efficient during training~\cite{lester2021power, li2021prefix, liu2021gpt, liu-etal-2022-p}. Continuous prompts extend prompt selection to the entire space of embeddings, including vector embeddings that do not correspond to any human-interpretable natural language tokens.  Hence, soft prompts are more expressive than discrete prompts. 

However, both deep prompts and shallow prompts assume a {\em static prompt / task-level prompt} for all samples within a task, neglecting the fact that samples might vary greatly, especially in the field of conversation generation. There are recent papers exploring possible {\em instance-specific prompts}.  For instance, Control-prefixes~\cite{clive2021control} generates attribute-level prompts for input labels, but its expressiveness is limited to four labels. IPL~\cite{jin2022instance} includes a look-up module to reweight prompt tokens before passing the updated embedding-only prompt into the transformer, but IPL updates all model parameters, which loses the efficiency benefits of prompting. IDPG~\cite{wu2022idpg} consumes inputs in a two-layer perceptron module to generate instance-dependent prompts in classification tasks rather than generation tasks. In addition, \cite{gu2021response} proposes DialogPrompt which performs instance-specific prompting for dialogue generation by conditioning the prompt on the entire dialogue history. However, their prompting module consists of GPT-2, which is a full-fledged language model, and the approach is as costly as storing an entire fine-tuned base model. Recent works Contrastive prefixes \cite{qian2022controllable} and Tailor \cite{yang2022tailor} both propose \textit{attribute-based prompts}, instead of instance-specific, to include either single-attribute or multi-attribute prompts into controlled text generation tasks, which reveal the powerful potential of controllability of continuous prompts.    

In contrast to previous work, we propose Controlled DialogPrompt for applying prompt-tuning in controlled dialogue generation, which optimizes prompts based on provided control codes rather than the previous conversation history and we further explore the controllability of prompts at the instance level. The size of the prompt encoder is strictly limited and we freeze the pretrained transformer during training in order to preserve memory efficiency. In addition, we would like to highlight that our work focuses more on open-ended text generation rather than natural language understanding, such as entailment, paraphrase detection, extractive QA, as seen in other parameter-efficient fine-tuning methods \cite{he2022hyperprompt,guo2021parameter, wu2022idpg}. We posit that generating high-quality text is a more challenging task that requires a more nuanced approach to prompt tuning.

\section{Controlled DialogPrompt}

In this section, we present Controlled DialogPrompt (Controlled DP) for dialogue generation, which is expected to provide attribute information such as the dialogue intention or the user’s persona within the prompt and steer the pretrained model efficiently. 

Soft Prompt-tuning~\cite{lester2021power,liu2021gpt} learns soft tokens for different tasks and then prepends them to the conversation context as well as control attributes. This approach yields a {\em static shallow} prompt since the soft tokens are static (i.e., fixed for a task) and shallow (only added as an input to the language model).

In contrast, Prefix-tuning proposes a more effective technique that adds soft tokens in the form of key-value pairs at every attention block of the transformer~\cite{li2021prefix,liu-etal-2022-p}. This allows the soft tokens to influence each stage of the language model and therefore it is referred to as a {\em static deep} prompt. 

Figure~\ref{fig:ControlledDP}(bottom right) shows our proposed controlled dialogue prompt (Deep version). Instead of training static soft tokens for the dialogue task, we train a lightweight prompt module that takes as input a control attribute, either an intention label or persona sentences, and outputs key-value pairs that are prepended to each layer of the language model. Since the soft token embeddings change depending on the control attribute, this corresponds to an {\em instance-specific} prompt.  For the shallow prompt (Figure~\ref{fig:ControlledDP} bottom left), we follow Soft Prompt-tuning which adds an additional trainable embedding layer to encode the attribute. For the deep prompt module, we consider two architectures: i) a simple multilayer perceptron (two fully connected layers of size 512 with tanh activation) applied to each token of the control attribute, and ii) a two-layer transformer decoder with embedding size of 256. The embedding size of each architecture was chosen to yield roughly the same number of parameters. This number of parameters is about 5\%-6\% of the number of parameters of the language model. For a given domain, training the prompt module is done as follows. An intention label or persona sentences are fed to the prompting module, which outputs key-value pairs added at each layer of the frozen pretrained dialogue system. Gradients to maximize the likelihood of response tokens are back-propagated through the dialogue system and prompting module, but only the weights of the prompting module are updated.

\begin{table*}
\centering
\begin{adjustbox}{width={\textwidth},totalheight={\textheight},keepaspectratio} %
\small\begin{tabular}{lrcccccccccc}
\hline
\textbf{Method} &  \textbf{$\phi \%$ } & \textbf{Controllability} & \multicolumn{2}{c}{\textbf{BLEU $\uparrow$}} & \multicolumn{2}{c}{\textbf{NIST $\uparrow$}} & \textbf{ROUGE-L $\uparrow$} & \textbf{METEOR $\uparrow$} & \multicolumn{2}{c}{\textbf{Dist $\uparrow$}} & \textbf{Entropy $\uparrow$} \\ \cline{4-5} \cline{6-7} \cline{10-11}
 &  & \textbf{Accuracy} & B-2 & B-4 & N-2 & N-4 &  &  & D-1 & D-2 & E-4 \\
\hline
Pretrained~\cite{zhang2020dialogpt} &  0$\%$ & 58.30$\%$ & 10.31$\%$& 1.73$\%$ & 0.18 & 0.18 & 19.43$\%$ & 7.30$\%$ & \color{red}\textbf{7.61$\%$} & \color{red}\textbf{40.00$\%$} & 10.03  \\
Fine-tuning &  100$\%$ & \color{red}\textbf{80.25$\%$} & \color{red}\textbf{21.03$\%$} & \color{red}\textbf{5.70$\%$} & 0.96  & 0.98 & \color{red}\textbf{34.38$\%$} & \color{red}\textbf{13.05$\%$} & 6.02$\%$& 34.51$\%$& 10.21 \\
Soft Prompt-tuning~\cite{lester2021power} & 0.008$\%$ & 70.51$\%$& 18.15$\%$ & 4.08$\%$ & 0.56 & 0.57 & 31.58$\%$& 11.46$\%$& 5.33$\%$& 30.82$\%$& 10.02 \\
Prefix-tuning~\cite{li2021prefix} & 3.1$\%$ & 75.02$\%$ & 19.94$\%$ & 5.12$\%$ & 0.91 & 0.93 & 33.29$\%$& 12.54$\%$& 5.59$\%$& 32.46$\%$& 10.17 \\
Controlled DialogPrompt (Embedding) & 0.001$\%$\tablefootnote{\label{paramdiff} Controlled DP (Embedding) involves training an embedding layer in a size of (prompt\_vocab\_size * base\_model\_n\_embd). In DialogAct control, we use only 4 labels, resulting in a size of 4 * 1280. In User's Persona, since there are many words in the corpus, we adopt the base model vocab size as the prompt vocab size and the embedding layer is 50257 * 1280. Therefore, the proportion of tunable parameters is higher in User's Persona Control.} 
& 69.06$\%$ & \color{blue}\textbf{20.11$\%$} &  4.91$\%$& 0.71 & 0.73 & 32.80$\%$& 12.19$\%$& 5.18$\%$& 30.07$\%$& 10.03 \\
\hline
Controlled DialogPrompt (MLP) & 3.1$\%$ & 78.36$\%$& 19.92$\%$& \color{blue}\textbf{5.43$\%$} & 0.98 & 1.01 & 33.12$\%$& 12.61$\%$& 5.71$\%$& 32.42$\%$& 10.20 \\ 
Controlled DialogPrompt (2-layer Transformer) & 3.3$\%$ & \color{blue}\textbf{78.58$\%$} & 19.86$\%$& 5.26$\%$&  \color{red}\textbf{1.01} &  \color{red}\textbf{1.04} &  \color{blue}\textbf{33.35$\%$} & \color{blue}\textbf{12.64$\%$} & \color{blue}\textbf{5.82$\%$} & \color{blue}\textbf{33.16$\%$} & \color{red}\textbf{10.23}  \\
\hline
\end{tabular}%
\end{adjustbox}
\caption{\textbf{DialogAct label} control performance under Dailydialog multi-reference evaluation. $\phi \%$ denotes the $\%$ of tunable parameters to the frozen-LM parameters required at training time. Red number is the best value in every metric on all methods. Blue number is the best value in every metric among prompting methods.
}
\label{dialogact-large}
\end{table*}

\begin{table*}
\centering
\begin{adjustbox}{width={\textwidth},totalheight={\textheight},keepaspectratio} %
\small\begin{tabular}{lrcccccccccc}
\hline
\textbf{Method} &  \textbf{$\phi \%$ } & \textbf{Controllability} & \multicolumn{2}{c}{\textbf{BLEU $\uparrow$}} & \multicolumn{2}{c}{\textbf{NIST $\uparrow$}} & \textbf{ROUGE-L $\uparrow$} & \textbf{METEOR $\uparrow$} & \multicolumn{2}{c}{\textbf{Dist $\uparrow$}} & \textbf{Entropy $\uparrow$} \\ \cline{4-5} \cline{6-7} \cline{10-11}
 &  & \textbf{Similarity} & B-2 & B-4 & N-2 & N-4 &  &  & D-1 & D-2 & E-4 \\
\hline
Pretrained~\cite{zhang2020dialogpt} &  0$\%$ & 51.40$\%$ & 1.63$\%$&  0.42$\%$ & 0.02 & 0.02 & 6.62$\%$ & 3.67$\%$ & 7.62$\%$ & 34.44$\%$ & 10.15  \\
Fine-tuning &  100$\%$ & \color{red}\textbf{75.21$\%$} & \color{red}\textbf{37.38$\%$} & \color{red}\textbf{25.77$\%$} &  \color{red}\textbf{5.80} & \color{red}\textbf{6.30} & \color{red}\textbf{27.71$\%$} & \color{red}\textbf{24.43$\%$} & \color{red}\textbf{7.93$\%$} & \color{red}\textbf{38.20$\%$}  & \color{red}\textbf{11.28}  \\
Soft Prompt-tuning~\cite{lester2021power} & 0.008$\%$ & 62.69$\%$& 18.01$\%$ & 9.50$\%$ & 2.72 & 2.87 & 16.53$\%$& 13.29$\%$& 6.77$\%$& 32.19$\%$ & 10.96 \\
Prefix-tuning~\cite{li2021prefix} & 6.2$\%$ & \color{blue}\textbf{66.89$\%$} & 27.18$\%$ & 16.73$\%$ & 4.35 & 4.63 & 21.38$\%$& 18.56$\%$ & 7.60$\%$& 36.88$\%$ & \color{blue}\textbf{11.25} \\
Controlled DialogPrompt (Embedding) & 8.3$\%$\textsuperscript{\ref{paramdiff}} & 61.16$\%$ & 13.01$\%$ & 5.12$\%$ & 1.89 & 1.96 & 14.84$\%$ & 10.28$\%$& 5.21$\%$& 26.45$\%$& 10.82 \\
\hline
Controlled DialogPrompt (MLP) & 6.2$\%$ &64.96$\%$ & 26.82$\%$& 17.09$\%$& 4.25& 4.54 & 21.40$\%$ & 18.47$\%$&7.85$\%$&37.58$\%$&11.18 \\
Controlled DialogPrompt (2-layer Transformer) & 5.0$\%$ & 66.34$\%$& \color{blue}\textbf{31.85$\%$} & \color{blue}\textbf{21.67$\%$} & \color{blue}\textbf{5.00} & \color{blue}\textbf{5.40} & \color{blue}\textbf{24.20$\%$} & \color{blue}\textbf{21.16$\%$} & \color{blue}\textbf{7.85$\%$} & \color{blue}\textbf{37.86$\%$} & 11.24  \\
\hline
\end{tabular}%
\end{adjustbox}
\caption{\textbf{User's Persona} control performance under FoCus validation dataset. $\phi \%$ denotes the $\%$ of tunable parameters to the frozen-LM parameters required at training time. Red number is the best value in every metric on all methods. Blue number is the best value in every metric among prompting methods.
}
\label{personality-large}
\end{table*}

\section{Experiments}

\subsection{Datasets and baseline models}

We evaluate the proposed method on two publicly available datasets: Dailydialog~\cite{lidailydialog} for label control and FoCus~\cite{jang2021call} for document control.  Dailydialog~\cite{lidailydialog} is a widely used daily conversation dataset that provides a dialogue act for every sentence that indicates the communication function of each utterance. There are 4 types of dialogue acts in total. FoCus\cite{jang2021call} is a new persona-grounded dataset that aims to provide informative answers based on the user’s persona about the geographical landmark. We provide the detailed dataset setups in Appendix~\ref{appendix:dataset}.  

To demonstrate better performance of Controlled DialogPrompt, we compare our model with other competitive prompt-tuning techniques. The backbone model is DialoGPT-Large~\cite{zhang2020dialogpt}. Details are provided in Appendix~\ref{appendix:baselinemodels}.

\subsection{Evaluation Methods}

We use both automatic evaluation metrics and human evaluation to measure the performance. 

\paragraph{Automated metrics}
For controllability, we follow \cite{du2021sidecontrol} to evaluate whether models can customize responses based on specified control attributes. Details about controllability measures are provided in Appendix~\ref{appendix:automated} Regarding response quality, we use n-gram based metrics such as BLEU (B-2, B-4) \cite{papineni2002bleu}, NIST (N-2, N-4) \cite{doddington2002automatic}, ROUGE-L \cite{lin2004rouge}, METEOR \cite{agarwal2007meteor} to evaluate fluency and adequacy and distinct n-gram distribution metrics such as Dist (D-1, D-2) \cite{li2016diversity} and Entropy (E-4) \cite{zhang2018generating} to measure the diversity of the response. 

\paragraph{Human Evaluation} 
Human evaluation on the other hand is used to measure consistency between dialogue context and response and attribute controllability. We adopt single-turn pairwise evaluations to prevent annotator bias in numerical score evaluation. Details on question settings and annotators are provided in Appendix~\ref{appendix:humaneval}

\section{Result and Analysis}

\subsection{DialogAct / Intention}

\begin{table}
\centering
\resizebox{0.5\textwidth}{!}{\begin{tabular}{lcc}
\hline
Methods & Attribute Relevancy  & Consistency \\
\hline
Controlled DP (Deep) & 30.7$\%$ & 32.0$\%$ \\
Soft Prompt-tuning & 20.0$\%$ & 20.0$\%$ \\
Neutral & 49.3$\%$ & 48.0$\%$\\
\hline
\\
\hline
Controlled DP (Deep) & 25.3$\%$ & 37.3$\%$\\
Prefix-tuning & 16.0$\%$ & 16.0$\%$\\
Neutral & 58.7$\%$ & 46.7$\%$\\
\hline
\\
\hline
Controlled DP (Deep) & 34.7$\%$ & 38.7$\%$\\
Controlled DP (Shallow) & 9.3$\%$ & 25.3$\%$ \\
Neutral & 56.0$\%$ & 36.0$\%$\\
\hline
\end{tabular} }
\caption{Human evaluation on Dailydialog dataset. "Controlled DP (Deep)" represents Controlled DialogPrompt with 2-layer transformer decoder as the prompt module. "Controlled DP (Shallow)" represents Controlled DialogPrompt on the embedding layer. “Neutral” means that there is no preference between the two answers according to the annotators.}
\label{dialogact-large-human}
\end{table}

Table~\ref{dialogact-large} summarizes the automatic evaluation results on the DialogAct label control task. Compared to static task prompts, instance-level controlled prompts achieve better performance consistently on both deep and shallow prompt levels. Since the controlled attribute is injected independently through the prompts, it does not affect the understanding and generation ability of the pretrained transformer. Both Controlled DP deep methods show higher controllability and response quality than Controlled DP embedding, in line with \cite{li2021prefix,liu-etal-2022-p,qin2021learning} indicating the expressiveness of deep prompts. Also, Controlled DP deep methods show performance close to fine-tuning and even outperform on some metrics such as NIST. This is because NIST is weighted-BLEU with higher weights on rarer words and fine-tuning tends to generate from a more limited vocabulary whereas Controlled DialogPrompt sometimes generates less frequent words and can attain a better NIST score. Human evaluation (Table~\ref{dialogact-large-human}) also shows that Controlled DP deep has a significantly higher winning rate than other prompting techniques on both control attribute relevancy and conversation consistency.

\subsection{User's Persona}

\begin{table}
\centering
\resizebox{0.5\textwidth}{!}{\begin{tabular}{lcc}
\hline
Methods & Persona Controllability  & Consistency \\
\hline
Controlled DP (Deep) & 41.3$\%$ & 44.0$\%$ \\
Soft Prompt-tuning & 5.3$\%$ & 13.3$\%$ \\
Neutral & 53.3$\%$ & 42.7$\%$\\
\hline
\\
\hline
Controlled DP (Deep) & 22.7$\%$ & 28.0$\%$\\
Prefix-tuning & 26.7$\%$ & 8.0$\%$\\
Neutral & 50.7$\%$ & 64.0$\%$\\
\hline
\\
\hline
Controlled DP (Deep) & 29.3$\%$ & 41.3$\%$\\
Controlled DP (Shallow) & 21.3$\%$ & 9.3$\%$ \\
Neutral & 49.3$\%$ & 49.3$\%$\\
\hline
\end{tabular} }
\caption{Human evaluation on Focus dataset. "Controlled DP (Deep)" represents Controlled DialogPrompt with 2-layer transformer decoder as the prompt module. "Controlled DP (Shallow)" represents Controlled DialogPrompt on the embedding layer. “Neutral” means that there is no preference between the two answers according to the annotators.}
\label{persona-large-human}
\end{table}

Table~\ref{personality-large} shows that our model displays advantages over other prompting methods in terms of response quality, which shows a promising sign that controlled DP can be adapted to more challenging document control scenarios. Note that the difference in BLEU-2 is more pronounced for Focus compared to DailyDialog, as Focus is more complicated and uses sentences as the attribute rather than labels. Although controlled DP methods perform slightly lower than Prefix-tuning on the similarity scores with given user's persona and Entropy-4 values, we find it to be highly consistent with the previous conversation history upon human evaluation (Table~\ref{persona-large-human}). Similar results are observed with FoCus \cite{jang2021call} where models with high generation abilities do not always ensure high grounding abilities. In addition, the difference between static/instance-specific deep prompts and static/instance-specific shallow prompts emphasizes the direct impact of deep prompts in complex tasks. Fine-tuning performs the best, but with approximately 20$X$ more tunable parameters.

\section{Conclusion and Future Work}

In summary, we presented a novel prompting technique, conditioned on a dialogue attribute (persona or intent), for controlled dialogue generation. The prompting module requires only 5\%-6\% of the total number of parameters, which allows the storage of several fined-tuned prompting modules for different dialogue generation tasks at a fraction of the cost of a full dialogue model. 

However, Controlled DialogPrompt currently studies conditioning on simple control attribute sentences like the user's persona and the work can be extended to more extensive and complex sentences such as background knowledge documents to further evaluate the controlled prompt's encoding capabilities. Additionally, combining multiple Controlled DialogPrompts on several control attributes and automatically triggering various dialogue skills is an interesting and unexplored direction.

\section*{Limitations}
In our current experiments, prompt-based methods are primarily storage-efficient or parameter-efficient solutions. Since these methods all require backpropagation to the bottom layer, the training time of prompt-based methods are closely resembles that of traditional fine-tuning approach. 

\section*{Acknowledgements}

This research was funded by Huawei Canada and the National Sciences and Engineering Research Council of Canada. Resources used in preparing this research at the University of Waterloo were provided by the province of Ontario and the government of Canada through CIFAR and
companies sponsoring the Vector Institute.

\bibliography{anthology,custom}

\begin{thebibliography}{36}
\expandafter\ifx\csname natexlab\endcsname\relax\def\natexlab#1{#1}\fi

\bibitem[{Agarwal and Lavie(2007)}]{agarwal2007meteor}
Abhaya Agarwal and Alon Lavie. 2007.
\newblock Meteor: An automatic metric for mt evaluation with high levels of
  correlation with human judgments.
\newblock \emph{Proceedings of WMT-08}.

\bibitem[{Ben-David et~al.(2021)Ben-David, Oved, and Reichart}]{ben2021pada}
Eyal Ben-David, Nadav Oved, and Roi Reichart. 2021.
\newblock Pada: A prompt-based autoregressive approach for adaptation to unseen
  domains.
\newblock \emph{arXiv preprint arXiv:2102.12206}.

\bibitem[{Brown et~al.(2020)Brown, Mann, Ryder, Subbiah, Kaplan, Dhariwal,
  Neelakantan, Shyam, Sastry, Askell et~al.}]{brown2020language}
Tom Brown, Benjamin Mann, Nick Ryder, Melanie Subbiah, Jared~D Kaplan, Prafulla
  Dhariwal, Arvind Neelakantan, Pranav Shyam, Girish Sastry, Amanda Askell,
  et~al. 2020.
\newblock Language models are few-shot learners.
\newblock \emph{Advances in neural information processing systems},
  33:1877--1901.

\bibitem[{Clive et~al.(2021)Clive, Cao, and Rei}]{clive2021control}
Jordan Clive, Kris Cao, and Marek Rei. 2021.
\newblock Control prefixes for text generation.
\newblock \emph{arXiv preprint arXiv:2110.08329}.

\bibitem[{Devlin et~al.(2019)Devlin, Chang, Lee, and
  Toutanova}]{devlin2019bert}
Jacob Devlin, Ming-Wei Chang, Kenton Lee, and Kristina Toutanova. 2019.
\newblock Bert: Pre-training of deep bidirectional transformers for language
  understanding.
\newblock In \emph{Proceedings of the 2019 Conference of the North American
  Chapter of the Association for Computational Linguistics: Human Language
  Technologies, Volume 1 (Long and Short Papers)}, pages 4171--4186.

\bibitem[{Doddington(2002)}]{doddington2002automatic}
George Doddington. 2002.
\newblock Automatic evaluation of machine translation quality using n-gram
  co-occurrence statistics.
\newblock In \emph{Proceedings of the second international conference on Human
  Language Technology Research}, pages 138--145.

\bibitem[{Du and Ji(2021)}]{du2021sidecontrol}
Wanyu Du and Yangfeng Ji. 2021.
\newblock Sidecontrol: Controlled open-domain dialogue generation via additive
  side networks.
\newblock In \emph{Findings of the Association for Computational Linguistics:
  EMNLP 2021}, pages 2175--2194.

\bibitem[{Gao et~al.(2021)Gao, Fisch, and Chen}]{gao2021making}
Tianyu Gao, Adam Fisch, and Danqi Chen. 2021.
\newblock Making pre-trained language models better few-shot learners.
\newblock In \emph{Proceedings of the 59th Annual Meeting of the Association
  for Computational Linguistics and the 11th International Joint Conference on
  Natural Language Processing (Volume 1: Long Papers)}, pages 3816--3830.

\bibitem[{Gu et~al.(2021)Gu, Yoo, and Lee}]{gu2021response}
Xiaodong Gu, Kang~Min Yoo, and Sang-Woo Lee. 2021.
\newblock Response generation with context-aware prompt learning.
\newblock \emph{arXiv preprint arXiv:2111.02643}.

\bibitem[{Guo et~al.(2021)Guo, Rush, and Kim}]{guo2021parameter}
Demi Guo, Alexander~M Rush, and Yoon Kim. 2021.
\newblock Parameter-efficient transfer learning with diff pruning.
\newblock In \emph{Proceedings of the 59th Annual Meeting of the Association
  for Computational Linguistics and the 11th International Joint Conference on
  Natural Language Processing (Volume 1: Long Papers)}, pages 4884--4896.

\bibitem[{Gupta et~al.(2019)Gupta, Mehri, Zhao, Pavel, Eskenazi, and
  Bigham}]{gupta2019investigating}
Prakhar Gupta, Shikib Mehri, Tiancheng Zhao, Amy Pavel, Maxine Eskenazi, and
  Jeffrey~P Bigham. 2019.
\newblock Investigating evaluation of open-domain dialogue systems with human
  generated multiple references.
\newblock In \emph{Proceedings of the 20th Annual SIGdial Meeting on Discourse
  and Dialogue}, pages 379--391.

\bibitem[{He et~al.(2022)He, Zheng, Tay, Gupta, Du, Aribandi, Zhao, Li, Chen,
  Metzler et~al.}]{he2022hyperprompt}
Yun He, Steven Zheng, Yi~Tay, Jai Gupta, Yu~Du, Vamsi Aribandi, Zhe Zhao,
  YaGuang Li, Zhao Chen, Donald Metzler, et~al. 2022.
\newblock Hyperprompt: Prompt-based task-conditioning of transformers.
\newblock In \emph{International Conference on Machine Learning}, pages
  8678--8690. PMLR.

\bibitem[{Jang et~al.(2021)Jang, Lim, Hur, Oh, Son, Lee, Shin, Kim, and
  Lim}]{jang2021call}
Yoonna Jang, Jungwoo Lim, Yuna Hur, Dongsuk Oh, Suhyune Son, Yeonsoo Lee,
  Donghoon Shin, Seungryong Kim, and Heuiseok Lim. 2021.
\newblock Call for customized conversation: Customized conversation grounding
  persona and knowledge.
\newblock \emph{arXiv preprint arXiv:2112.08619}.

\bibitem[{Jiang et~al.(2020)Jiang, Xu, Araki, and Neubig}]{jiang2020can}
Zhengbao Jiang, Frank~F Xu, Jun Araki, and Graham Neubig. 2020.
\newblock How can we know what language models know?
\newblock \emph{Transactions of the Association for Computational Linguistics},
  8:423--438.

\bibitem[{Jin et~al.(2022)Jin, Lu, Zhang, and Zong}]{jin2022instance}
Feihu Jin, Jinliang Lu, Jiajun Zhang, and Chengqing Zong. 2022.
\newblock Instance-aware prompt learning for language understanding and
  generation.
\newblock \emph{arXiv preprint arXiv:2201.07126}.

\bibitem[{Lester et~al.(2021)Lester, Al-Rfou, and Constant}]{lester2021power}
Brian Lester, Rami Al-Rfou, and Noah Constant. 2021.
\newblock The power of scale for parameter-efficient prompt tuning.
\newblock In \emph{Proceedings of the 2021 Conference on Empirical Methods in
  Natural Language Processing}, pages 3045--3059.

\bibitem[{Li et~al.(2016)Li, Galley, Brockett, Gao, and
  Dolan}]{li2016diversity}
Jiwei Li, Michel Galley, Chris Brockett, Jianfeng Gao, and William~B Dolan.
  2016.
\newblock A diversity-promoting objective function for neural conversation
  models.
\newblock In \emph{Proceedings of the 2016 Conference of the North American
  Chapter of the Association for Computational Linguistics: Human Language
  Technologies}, pages 110--119.

\bibitem[{Li et~al.(2019)Li, Weston, and Roller}]{li2019acute}
Margaret Li, Jason Weston, and Stephen Roller. 2019.
\newblock Acute-eval: Improved dialogue evaluation with optimized questions and
  multi-turn comparisons.
\newblock \emph{arXiv preprint arXiv:1909.03087}.

\bibitem[{Li and Liang(2021)}]{li2021prefix}
Xiang~Lisa Li and Percy Liang. 2021.
\newblock Prefix-tuning: Optimizing continuous prompts for generation.
\newblock In \emph{Proceedings of the 59th Annual Meeting of the Association
  for Computational Linguistics and the 11th International Joint Conference on
  Natural Language Processing (Volume 1: Long Papers)}, pages 4582--4597.

\bibitem[{Li et~al.()Li, Su, Shen, Li, Cao, and Niu}]{lidailydialog}
Yanran Li, Hui Su, Xiaoyu Shen, Wenjie Li, Ziqiang Cao, and Shuzi Niu.
\newblock Dailydialog: A manually labelled multi-turn dialogue dataset.

\bibitem[{Lin(2004)}]{lin2004rouge}
Chin-Yew Lin. 2004.
\newblock Rouge: A package for automatic evaluation of summaries.
\newblock In \emph{Text summarization branches out}, pages 74--81.

\bibitem[{Liu et~al.(2022)Liu, Ji, Fu, Tam, Du, Yang, and
  Tang}]{liu-etal-2022-p}
Xiao Liu, Kaixuan Ji, Yicheng Fu, Weng Tam, Zhengxiao Du, Zhilin Yang, and Jie
  Tang. 2022.
\newblock \href {https://doi.org/10.18653/v1/2022.acl-short.8} {{P}-tuning:
  Prompt tuning can be comparable to fine-tuning across scales and tasks}.
\newblock In \emph{Proceedings of the 60th Annual Meeting of the Association
  for Computational Linguistics (Volume 2: Short Papers)}, pages 61--68,
  Dublin, Ireland. Association for Computational Linguistics.

\bibitem[{Liu et~al.(2021)Liu, Zheng, Du, Ding, Qian, Yang, and
  Tang}]{liu2021gpt}
Xiao Liu, Yanan Zheng, Zhengxiao Du, Ming Ding, Yujie Qian, Zhilin Yang, and
  Jie Tang. 2021.
\newblock Gpt understands, too.
\newblock \emph{arXiv preprint arXiv:2103.10385}.

\bibitem[{Papineni et~al.(2002)Papineni, Roukos, Ward, and
  Zhu}]{papineni2002bleu}
Kishore Papineni, Salim Roukos, Todd Ward, and Wei-Jing Zhu. 2002.
\newblock Bleu: a method for automatic evaluation of machine translation.
\newblock In \emph{Proceedings of the 40th annual meeting of the Association
  for Computational Linguistics}, pages 311--318.

\bibitem[{Qian et~al.(2022)Qian, Dong, Shen, Wei, and
  Chen}]{qian2022controllable}
Jing Qian, Li~Dong, Yelong Shen, Furu Wei, and Weizhu Chen. 2022.
\newblock Controllable natural language generation with contrastive prefixes.
\newblock In \emph{Findings of the Association for Computational Linguistics:
  ACL 2022}, pages 2912--2924.

\bibitem[{Qin and Eisner(2021)}]{qin2021learning}
Guanghui Qin and Jason Eisner. 2021.
\newblock Learning how to ask: Querying lms with mixtures of soft prompts.
\newblock In \emph{Proceedings of the 2021 Conference of the North American
  Chapter of the Association for Computational Linguistics: Human Language
  Technologies}, pages 5203--5212.

\bibitem[{Radford et~al.()Radford, Narasimhan, Salimans, Sutskever
  et~al.}]{radford2018improving}
Alec Radford, Karthik Narasimhan, Tim Salimans, Ilya Sutskever, et~al.
\newblock Improving language understanding by generative pre-training.

\bibitem[{Roller et~al.(2021)Roller, Dinan, Goyal, Ju, Williamson, Liu, Xu,
  Ott, Smith, Boureau et~al.}]{roller2021recipes}
Stephen Roller, Emily Dinan, Naman Goyal, Da~Ju, Mary Williamson, Yinhan Liu,
  Jing Xu, Myle Ott, Eric~Michael Smith, Y-Lan Boureau, et~al. 2021.
\newblock Recipes for building an open-domain chatbot.
\newblock In \emph{Proceedings of the 16th Conference of the European Chapter
  of the Association for Computational Linguistics: Main Volume}, pages
  300--325.

\bibitem[{Schick and Sch{\"u}tze(2021{\natexlab{a}})}]{schick2021exploiting}
Timo Schick and Hinrich Sch{\"u}tze. 2021{\natexlab{a}}.
\newblock Exploiting cloze-questions for few-shot text classification and
  natural language inference.
\newblock In \emph{Proceedings of the 16th Conference of the European Chapter
  of the Association for Computational Linguistics: Main Volume}, pages
  255--269.

\bibitem[{Schick and Sch{\"u}tze(2021{\natexlab{b}})}]{schick2021s}
Timo Schick and Hinrich Sch{\"u}tze. 2021{\natexlab{b}}.
\newblock It’s not just size that matters: Small language models are also
  few-shot learners.
\newblock In \emph{Proceedings of the 2021 Conference of the North American
  Chapter of the Association for Computational Linguistics: Human Language
  Technologies}, pages 2339--2352.

\bibitem[{Shin et~al.(2020)Shin, Razeghi, Logan~IV, Wallace, and
  Singh}]{shin2020autoprompt}
Taylor Shin, Yasaman Razeghi, Robert~L Logan~IV, Eric Wallace, and Sameer
  Singh. 2020.
\newblock Autoprompt: Eliciting knowledge from language models with
  automatically generated prompts.
\newblock In \emph{Proceedings of the 2020 Conference on Empirical Methods in
  Natural Language Processing (EMNLP)}, pages 4222--4235.

\bibitem[{Wu et~al.(2022)Wu, Wang, Gu, Hou, Dong, Vydiswaran, and
  Ma}]{wu2022idpg}
Zhuofeng Wu, Sinong Wang, Jiatao Gu, Rui Hou, Yuxiao Dong, VG~Vydiswaran, and
  Hao Ma. 2022.
\newblock Idpg: An instance-dependent prompt generation method.
\newblock \emph{arXiv preprint arXiv:2204.04497}.

\bibitem[{Yang et~al.(2022)Yang, Liu, Lei, Yang, Xue, Chen, and
  Xie}]{yang2022tailor}
Kexin Yang, Dayiheng Liu, Wenqiang Lei, Baosong Yang, Mingfeng Xue, Boxing
  Chen, and Jun Xie. 2022.
\newblock Tailor: A prompt-based approach to attribute-based controlled text
  generation.
\newblock \emph{arXiv preprint arXiv:2204.13362}.

\bibitem[{Zhang et~al.(2018)Zhang, Galley, Gao, Gan, Li, Brockett, and
  Dolan}]{zhang2018generating}
Yizhe Zhang, Michel Galley, Jianfeng Gao, Zhe Gan, Xiujun Li, Chris Brockett,
  and Bill Dolan. 2018.
\newblock Generating informative and diverse conversational responses via
  adversarial information maximization.
\newblock \emph{Advances in Neural Information Processing Systems}, 31.

\bibitem[{Zhang et~al.(2020)Zhang, Sun, Galley, Chen, Brockett, Gao, Gao, Liu,
  and Dolan}]{zhang2020dialogpt}
Yizhe Zhang, Siqi Sun, Michel Galley, Yen-Chun Chen, Chris Brockett, Xiang Gao,
  Jianfeng Gao, Jingjing Liu, and William~B Dolan. 2020.
\newblock Dialogpt: Large-scale generative pre-training for conversational
  response generation.
\newblock In \emph{Proceedings of the 58th Annual Meeting of the Association
  for Computational Linguistics: System Demonstrations}, pages 270--278.

\bibitem[{Zhao et~al.(2021)Zhao, Wallace, Feng, Klein, and
  Singh}]{zhao2021calibrate}
Tony~Z Zhao, Eric Wallace, Shi Feng, Dan Klein, and Sameer Singh. 2021.
\newblock Calibrate before use: Improving few-shot performance of language
  models.
\newblock \emph{arXiv preprint arXiv:2102.09690}.

\end{thebibliography}
\bibliographystyle{acl_natbib}

\appendix

\section{Experimental Setups}

\subsection{Datasets}
\label{appendix:dataset}

\subsubsection{Label control} 
Dailydialog~\cite{lidailydialog} is a widely used daily conversation dataset that provides a dialogue act for every sentence. Dialogue acts indicate the communication function of each utterance and there are 4 types of dialogue acts: inform, questions, directives, and commissives. We follow the standard split of the original Dailydialog dataset, limit the conversation context to a maximum of four sentences, and remove any sentence that has more than 25 words to maintain computation efficiency. As a result, we obtain 61,669 training samples, 5769 validation samples, and 5453 testing samples. 

We additionally use the Dailydialog multi-reference dataset from \cite{gupta2019investigating} during metrics computation to mitigate the one-to-many possible response problem.

\subsubsection{Document control}
FoCus\cite{jang2021call} is a persona-grounded dataset. Unlike DailyDialog, FoCus aims to build a dialogue agent that provides informative answers based on the user’s persona about the geographical landmark; therefore, it is more content-rich and challenging. The selected knowledge candidate sentence is prepended to the conversation and regarded as part of the input. 

The input to the base model has the template: "\textit{Knowledge: [Selected knowledge sentence] Conversation: [Previous utterances]}”. The persona sentences are given as the input to the prompt encoder. In fine-tuning (no prompt encoder) and static prompt methods (the prompt encoder does not take attribute information), the persona sentences are concatenated together with the knowledge and previous utterances and form the input to base model as “\textit{Knowledge: [Selected knowledge sentence] Persona: [User’s Personas] Conversation: [Previous utterances]}”

Since the grounded answer of the test set has not been released, we shuffle and split the original training set to construct our training samples and validation samples (70\% training and 30\% validation) and the original validation set as our testing samples. We further restrict conversation context to at most three sentences because the bot’s utterances are much longer than human’s utterances. In total, we have 49,198 samples for training, 21,134 samples for validation, and 5,639 samples for testing.  

\subsection{Baseline models}
\label{appendix:baselinemodels}
To demonstrate better performance of Controlled DialogPrompt, we compare our model with other competitive prompt-tuning techniques. 

\begin{itemize}
\item \textbf{Pretrained DialoGPT} \cite{zhang2020dialogpt}: DialoGPT-large  has shown its superiority for a wide range of open-domain dialogue generation tasks by pretraining on a massive corpus. 

\item \textbf{Fine-tuning}: Fine-tuning, though memory-consuming, is the most straightforward and prevalent adaptation technique to downstream tasks. Fine-tuning has been considered as the benchmark for all light-weight fine-tuning methods including prompt-tuning.

\item \textbf{Soft Prompt-tuning (static shallow prompt)} \cite{lester2021power}: The method applies a static task prompt to the embedding of every input. We experiment with different lengths (length 10 and length 50) of the static shallow prompt and use the better length 50. 

\item \textbf{Prefix-tuning (static deep prompt)} \cite{li2021prefix}: Prefix prompts are added to every layer during computation. We experiment with different lengths (length 10 and length 50) and we report the better prompt result with length 10.

\item \textbf{Controlled DP - Embedding (instance-specific shallow prompt)}: The shallow version of our method with controlled prompts added only in the embedding layer. It is used to demonstrate the expressiveness of the deep Controlled DialogPrompt.

\item \textbf{Controlled DP - MLP / 2-layer Transformer (instance-specific deep prompt)}: We explore different prompt encoder structures, among which MLP prompt encoder shares the frozen pretrained transformer embedding layer to reduce tunable parameters.
\end{itemize}

During our experiments, we utilize DialoGPT-large as the frozen backbone model and train all models on two Nvidia V100 32G GPUs. We train models for 10 epochs with training batch size 2 per GPU and learning rate of 1e-4 except for fine-tuning, which is set to 5e-5 in the FoCus dataset and 1e-5 in the Dailydialog dataset. Models that achieve the lowest validation losses are saved during the training. We perform optimization with the AdamW optimizer with maximum gradient clipping set to 1. For decoding, we choose top-k sampling provided in Huggingface where k=10 and temperature T=0.9. The result is generated with random seed=42.

\section{Evaluation Methods}

\subsection{Automated metrics} 
\label{appendix:automated}

For controllability, we follow \cite{du2021sidecontrol} to evaluate whether models can customize responses based on specified control attributes. (1) For label control, we fine tune an independent BERT classifier \cite{devlin2019bert} which can take a sentence and predict its dialogue intention. We train the classifier on the same training set and achieve 83.23\% accuracy on the test set. (2) For document control, we also compute the cosine similarity between the Glove embedding of the generated responses and grounded persona documents. As FoCus dataset contains human-annotated labels for used persona sentences, only those that are actually used are evaluated. Detailed training information is provided in \cite{du2021sidecontrol}.

Regarding response quality, we utilize different variants of n-gram based metrics such as BLEU (B-2, B-4) \cite{papineni2002bleu}, NIST (N-2, N-4) \cite{doddington2002automatic}, ROUGE-L \cite{lin2004rouge}, METEOR \cite{agarwal2007meteor} to evaluate fluency and adequacy and distinct n-gram distribution metrics such as Dist (D-1, D-2) \cite{li2016diversity} and Entropy (E-4) \cite{zhang2018generating} to measure the diversity of the response. We follow the metrics setting in \cite{zhang2020dialogpt}.

\subsection{Human Evaluation}
\label{appendix:humaneval}

Human evaluation on the other hand is used to measure consistency between dialogue context and response and attribute controllability. Similar to ACUTE-Eval in \cite{li2019acute, roller2021recipes}, we adopt single-turn pairwise evaluations to prevent annotator bias in numerical score evaluation. We compare Controlled DialogPrompt with every other prompt-tuning methods, covering static shallow prompt, static deep prompt and instance-specific shallow prompt. In each comparison group, there are two questions designed separately to assess response’s dialogact/personality controllability as well as consistency to the previous conversation context. For dialogact controllability, we have the question: \textit{Which response do you think is more related to the given dialog act (intention)?}. For personality controllability, we set the question as \textit{Which response do you think is more related to the personality?}. For the consistency to the previous conversation context, we set the question as \textit{Which response do you think is more consistent to the above conversation context?} We sample 15 conversations from each comparison group and there are 5 conversations overlapped across different groups. Annotators are industrial NLP researchers and NLP graduate students. We collected 900 annotations in total.

\end{document}